\documentclass[10pt,twocolumn,letterpaper]{article}

\usepackage{iccv}
\usepackage{times}
\usepackage{epsfig}
\usepackage{graphicx}
\usepackage{amsmath}
\usepackage{amssymb}
\usepackage{bbold}
\usepackage{bm}
\usepackage{float}
\usepackage{graphicx}
\usepackage{subfigure}
\usepackage{diagbox}
\usepackage{multirow}
\usepackage{color}

\usepackage{array}
\newcolumntype{C}[1]{>{\centering\let\newline\\\arraybackslash\hspace{0pt}}m{#1}}
\usepackage{booktabs}

\newcommand{\repeatthanks}{\textsuperscript{\thefootnote}}

\usepackage[breaklinks=true,bookmarks=false]{hyperref}

\iccvfinalcopy 

\def\iccvPaperID{1045} 

\ificcvfinal\fi

\begin{document}

\title{Explicit Shape Encoding for Real-Time Instance Segmentation}

\author{Wenqiang Xu\thanks{these two authors have equal contributions.} \and Haiyang Wang\repeatthanks \and Fubo Qi \and Cewu Lu\thanks{Cewu Lu is the corresponding author.}\thanks{Cewu Lu is a member of MoE Key Lab of Artificial Intelligence, AI Institute, Shanghai Jiao Tong University, and SJTU-SenseTime AI lab.}\\
Department of Computer Science and Engineering\\
Shanghai Jiao Tong University\\
{\tt\small \{vinjohn, wanghaiyang, 727749815, lucewu\}@sjtu.edu.cn}
}

\maketitle
\ificcvfinal\thispagestyle{empty}\fi

\begin{abstract}
   In this paper, we propose a novel top-down instance segmentation framework based on explicit shape encoding, named \textbf{ESE-Seg}. It largely reduces the computational consumption of the instance segmentation by explicitly decoding the multiple object shapes with tensor operations, thus performs the instance segmentation at almost the same speed as the object detection. ESE-Seg is based on a novel shape signature Inner-center Radius (IR), Chebyshev polynomial fitting and the strong modern object detectors. ESE-Seg with YOLOv3 outperforms the Mask R-CNN on Pascal VOC 2012 at mAP$^r$@0.5 while 7 times faster. Code and trained models are released at {\color{blue} \href{https://github.com/WenqiangX/ese\_seg}{https://github.com/WenqiangX/ese\_seg}}.
\end{abstract}

\section{Introduction}
\begin{figure}[t!]
\begin{center}
   \includegraphics[width=1\linewidth]{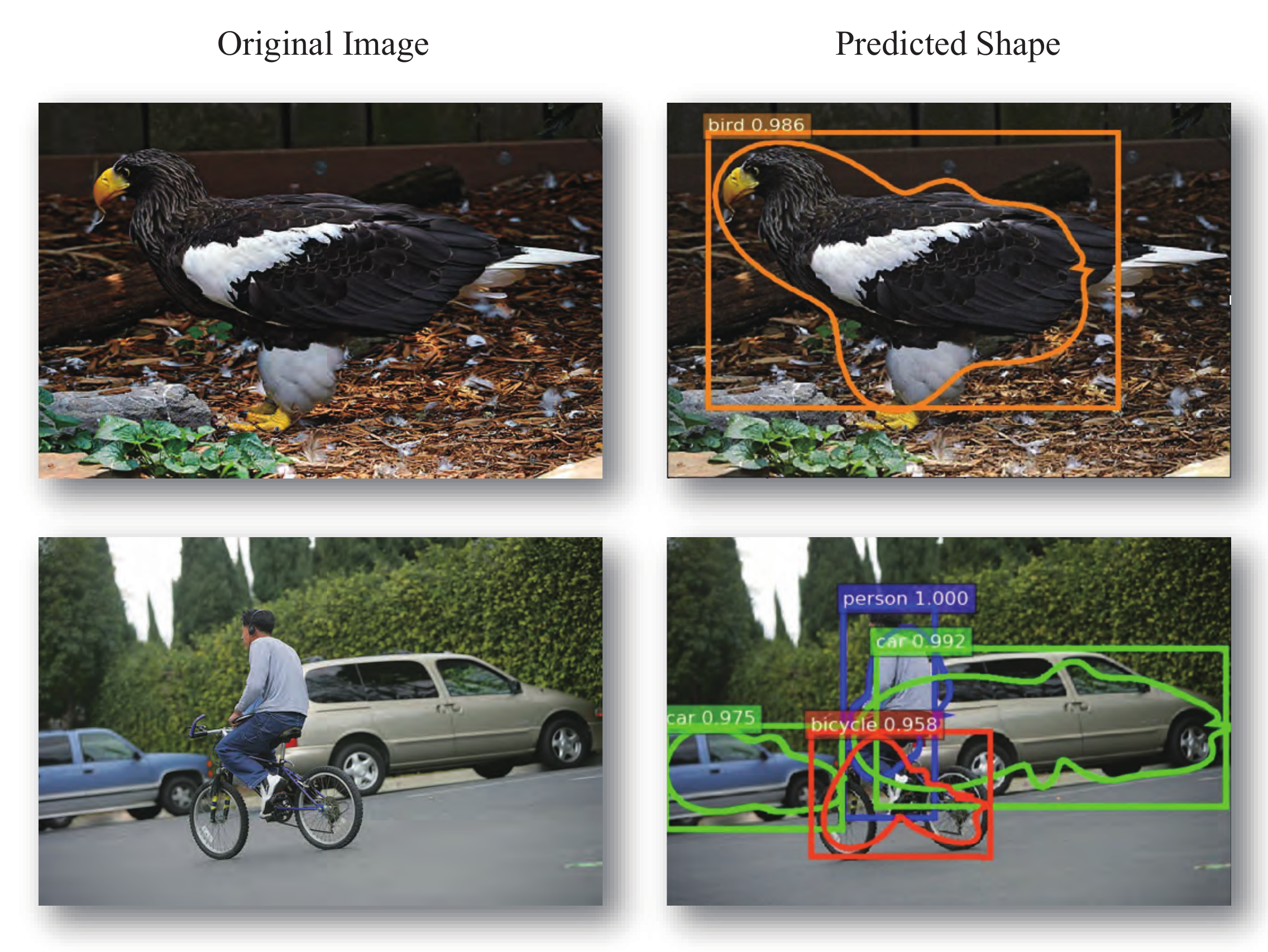}
\end{center}
   \caption{ESE-Seg learns to estimate the shapes of the detected objects, it can be simutaneously obtained along with the bounding boxes.}
\label{fig:representative_fig}
\end{figure}

Instance segmentation is a fundamental task in the computer vision, which is important for many real-world applications such as autonomous driving, robot manipulation. As the task seeks to predict both the object location and the shape, the methods for the instance segmentation are generally not as efficient as the object detection frameworks. Forwarding each object instance through an upsampling network to obtain the instance shape, as mainstream instance segmentation frameworks do \cite{he2017maskrcnn, liu2018path,dai2016instance,li2017fully}, is quite computation-consuming, especially when compared with the object detection which only needs to regress the bounding box, \ie a 4D vector for each object. Thus, if the network can also regress the object shape to a short vector, and decode the vector to the shape (see Fig. \ref{fig:representative_fig}) in a simple way just like the bounding box, it can make the instance segmentation reach \textit{almost} equal computational efficiency to the object detection. To achieve this goal, we propose a novel instance segmentation framework based on \textbf{E}xplicit \textbf{S}hape \textbf{E}ncoding and modern object detectors, named \textbf{ESE-Seg}.

Shape encoding is originally developed for instance retrieval \cite{zhang2002comparative, KIM200087,YOUNG1974357}, which encodes the object to a shape vector. Recently, a number of works encode the shape implicitly \cite{wang_retrieval4, wang_retrieval1,wang_retrieval3}, which is to project the shape content to a latent vector, typically through a black-box design such as deep CNN. Thus the decoding procedure under this approach should be also put through a network, which requires several forwarding for multiple instances, and causes large computation. In pursuit of fast decoding, we employ an explicit shape encoding that involves only simple numeric transformations.

However, designing a satisfactory explicit shape encoding method is non-trivial. Concerning the CNN training, as it is known to regress with uncertainties, a preferred shape vector should be relatively \textbf{short} but contains sufficient information, \textbf{robust} to the noise, and \textbf{efficiently decodable} to reconstruct the shape. In this paper, we propose a contour-based shape signature to meet these requirements. A novel ``Inner-center Radius'' (IR) shape signature for instance shape representation is introduced. The IR first locates an inner-center inside the object segment, and based on this inner-center, it transforms the contour points to polar coordinates. That is, we can form a function of radius $\bm{f}(\theta)$ along the contour with respect to angle $\theta$. To make the shape vector even shorter and more robust, we apply the Chebyshev polynomials to conduct the function approximation on $\bm{f}(\theta)$. As such, the IR signature is represented by a small number of coefficients with small error, and these coefficients are the shape vector to be predicted. Additionally, we also in-depth discuss about the comparison with other shape signature designs. Conventional object detector (\eg YOLOv3 \cite{yolov3}) is used to regress the shape vector, along with 4D bounding box vector. To note that our shape decoding can be implemented by simple tensor operations (multiplication and addition) which are extremely fast.

The ESE-Seg itself is independent of all the bounding box-based object detection frameworks \cite{faster_rcnn,R-FCN, girshick2015fast, retinanet,ssd}. We demonstrate the generality on Faster R-CNN \cite{faster_rcnn}, RetinaNet \cite{retinanet}, YOLO \cite{yolo} and YOLOv3-tiny \cite{yolov3} and evaluate our ESE-Seg on standard public datasets, namely Pascal VOC \cite{Everingham10} and COCO \cite{coco}. Our method achieves 69.3 mAP$^r$, 48.7 mAP respectively with IOU threshold 0.5. The score is better than Mask R-CNN \cite{he2017maskrcnn} on Pascal VOC 2012, and is competitive to the performance on COCO. It is decent considering it is 7 times faster than Mask R-CNN with the same backbone ResNet-50 \cite{he2016deep}. The speed can be even faster at $\sim$130fps on GTX 1080Ti when the base detector changes to YOLOv3-tiny, while the mAP$^r@0.5$ remains 53.2\% on the Pascal VOC. It is noteworthy, ESE-Seg speeds up the instance segmentation not depending on the model acceleration techniques \cite{mobilenets,shufflenet}, but relying on a new mechanism that cut down shape prediction after object detection.

\textbf{Contributions.} We propose an explicit shape encoding based instance segmentation framework, ESE-Seg. It is a top-down approach but reconstructs the shapes for multiple instances in one pass, thus greatly reduces the computational consumption, and makes the instance segmentation reach the speed of the object detection with no model acceleration techniques involved.

\section{Related Work}
\label{sec:rel}
\paragraph{Explicit v.s Implicit Shape Representation}
A previous work with similar ideology has been done by Jetley \etal \cite{wang_straight_to_shape_realtime_detect}. They took the implicit shape representation path by first training an autoencoder on object binary mask. The encoded shape vector is decoded to shape mask through the decoder component. In the implementation, they adopted the YOLO \cite{yolo} to regress the bounding box and the shape vector for each detected object. The YOLO structure can thus be viewed as both detector and encoder. The encoded vector from YOLO is then decoded by the pre-trained denoising autoencoder. The major differences between our work and theirs:
\begin{itemize}
   \item Explicit shape representation is typically based on the contour, while implicit shape representation is typically based on the mask.
   \item Explicit shape representation requires no additional decoder network training. Parallelizing the decoding process for all objects in the images, which is hard for network structured decoder, can be easily achieved by the explicit shape encoding. As a matter of fact, implicit decoding requires multiple passes for multiple objects, one for each, while explicit decoding can obtain all the shapes in one pass.
   \item The input for training autoencoder and training YOLO (viewed as an encoder) is quite different (object scales, color pattern), which may cause trouble for the decoder, since the decoder is not further optimized with YOLO training. Such an issue does not exist for explicit shape representation.
\end{itemize}
In addition to our proposed IR shape signature, there exist various methods to represent the shape, to name a few, centroid radius, complex coordinates, cumulative angle \cite{davies1997machine,van1991contour,zhang2002comparative} \etc. While such methods sample the shape related feature along the contour, only a few of them can be decoded to reconstruct the shape.

\paragraph{Object detection} Object detection is a richly studied field. Object detection frameworks with CNN can be roughly divided into two categories, one-stage and multi-stage. Two-stage detection scheme is a classic multi-stage scheme, which typically learns an RPN to sample region proposals and then refine the detection with roi pooling or its variations, the representative works are Faster R-CNN \cite{faster_rcnn}, R-FCN \cite{R-FCN}. Recently, some works extend the two-stage to multi-stage in a cascade form \cite{cai18cascadercnn}. On the other hand, one-stage detectors divide the input image to size-fixed grid cells and parallelize the detection on each cell with fully convolutional operations, the representative networks are SSD \cite{ssd}, YOLO \cite{yolo}, RetinaNet \cite{retinanet}. Recently, point-based detections are proposed, CornerNet \cite{law2018cornernet} directly detects the upper-left and bottom-right points, which is a one-stage detector. Grid R-CNN \cite{Lu2018GridR} regresses 9 points to construct the bounding box, which is a two-stage detector.
\begin{figure*}[t!]
\begin{center}
   \includegraphics[width=1\linewidth]{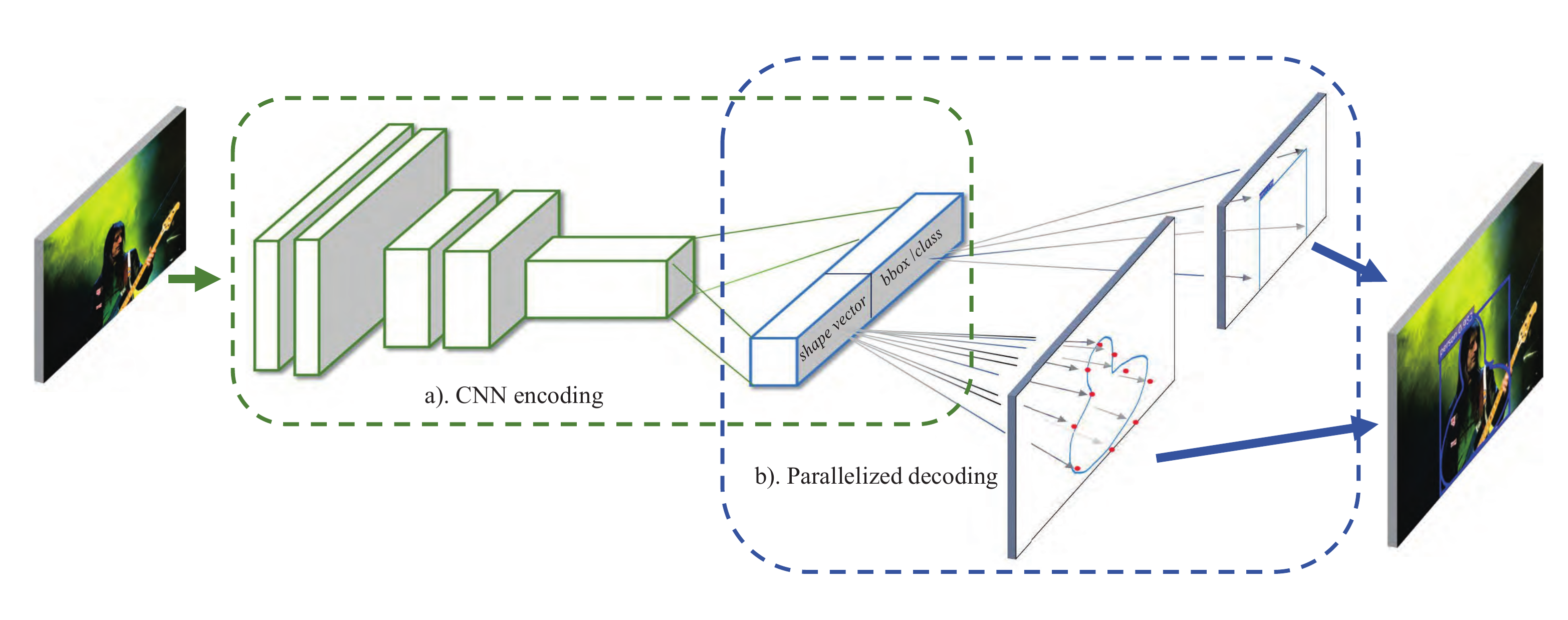}
\end{center}
   \caption{The pipeline of the shape detection, regression and reconstruction.}
\label{fig:overall}
\end{figure*}
Our method is compatible with all the bounding box-based detection networks. We experiment with Faster R-CNN, YOLO, YOLOv3, and RetinaNet to prove the generality. See Table \ref{exp:base}. However, it is not compatible with the point-based detector, as the shape (bounding box) in this setting is not parametrized.

\paragraph{Instance Segmentation}
Instance segmentation requires not only to locate the object instance but also to delineate the shape. The mainstream methods can be roughly divided to top-down \cite{he2017maskrcnn, liu2018path,dai2016instance,li2017fully, pang2019deep, novotny2018semi, chen2019hybrid} or bottom-up \cite{Newell:2017:AEE:3294771.3294988,Weinberger:2009:DML:1577069.1577078} approaches. Ours belongs to the top-down line. The top-down approaches such as MNC \cite{dai2016instance}, FCIS \cite{li2017fully}, Mask R-CNN \cite{he2017maskrcnn} are generally slowed down when the object number in an image is large, as they predict the instance mask in sequence. On the contrary, our ESE-Seg alleviates the cumbersome computation by regressing the object shapes to short vectors and decoding them simultaneously. It is also the first top-down instance segmentation framework which is not affected by the instance number in the images with respect to the inference time.
Besides, the works on augmenting the performance of instance segmentation frameworks through data augmentation \cite{Fang_2019_InstaBoost, Xu_2018_ECCV}, scale normalization \cite{sniper2018} can be easily integrated to our system.

\section{Method}
\subsection{Overview}
\label{sec:overview}
We propose an explicit shape encoding based detection to solve the instance segmentation. It predicts all the instance segments in one forwarding pass, which can reach equal efficiency as object detection solver. Given an object instance segment, we parametrize the contour with a novel shape signature ``Inner-center Radius'' (IR) (Sec. \ref{sec:descriptor}). The Chebyshev polynomials are used to approximate the shape signature vector with a small number of coefficients (Sec. \ref{sec:fitting}). Those coefficients are served as the shape descriptor, and the network will learn to regress it. (Sec. \ref{sec:learning}). Finally, we describe how to decode the shape descriptor under the ordinary object detection framework by simple tensor operations. (Sec. \ref{sec:decode}). The overall pipeline is shown in Fig. \ref{fig:overall}.

\paragraph{The Advantage of Explicit Shape Encoding} In object detection system (\eg YOLOv3), the network regresses the bounding boxes (\ie 4D vectors) and the bounding box is decoded by tensor operations, which is light to process and easy to parallelize. By contrast, conventional instance segmentation (\eg Mask R-CNN) requires an add-on network structure to compute the object shape. The decoding/upsampling forwarding involves a large number of parameters, which is heavy to load in parallel for multiple instances. This is why instance segmentation is normally much slower than object detection. Therefore, if we also regress the object shape into short vectors directly, the instance shape decoding can be achieved by fast tensor operations (multiplication and addition) in a similar way. Thus the instance segmentation can reach the speed of object detection.

\subsection{Shape Signature}

\subsubsection{Inner-center Radius Shape Signature}
\label{sec:descriptor}
In this section, we will describe the design of the ``inner-center radius'' shape signature and compare it to previously proposed shape signatures.

The construction of the ``inner-center radius'' contains two steps: First, locate an Inner center point inside the object segment as the origin point to build the polar coordinate system. Second, sampling the contour points according to the angle $\theta$. This signature is translation-invariant and scale-invariant after normalized.

\begin{figure}[ht!]
\begin{center}
   \includegraphics[width=1\linewidth]{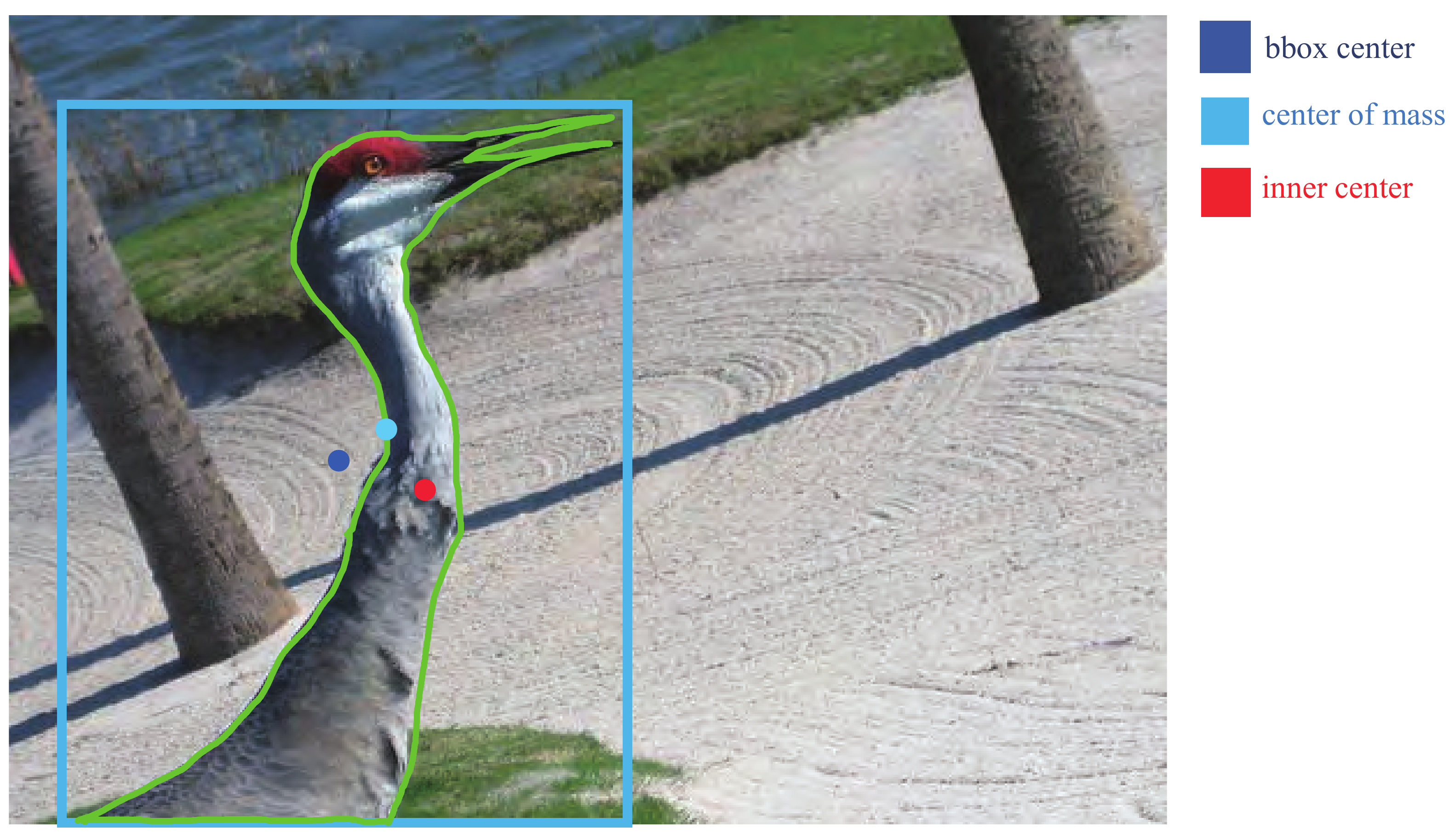}
\end{center}
   \caption{The center points of an object. As we can see, bounding box center and the center of mass cannot guarantee to be inside an object.}
\label{fig:center}
\end{figure}

\paragraph{Inner center} The inner-center point is defined by the most far-way point from the contour, which can be obtained through distance transform \cite{maurer2003linear}. To note, some commonly used center such as the center of mass, the center of the bounding box cannot guarantee to be inside the object. See Fig. \ref{fig:center}.

\begin{figure}[ht!]
\begin{center}
   \includegraphics[width=1\linewidth]{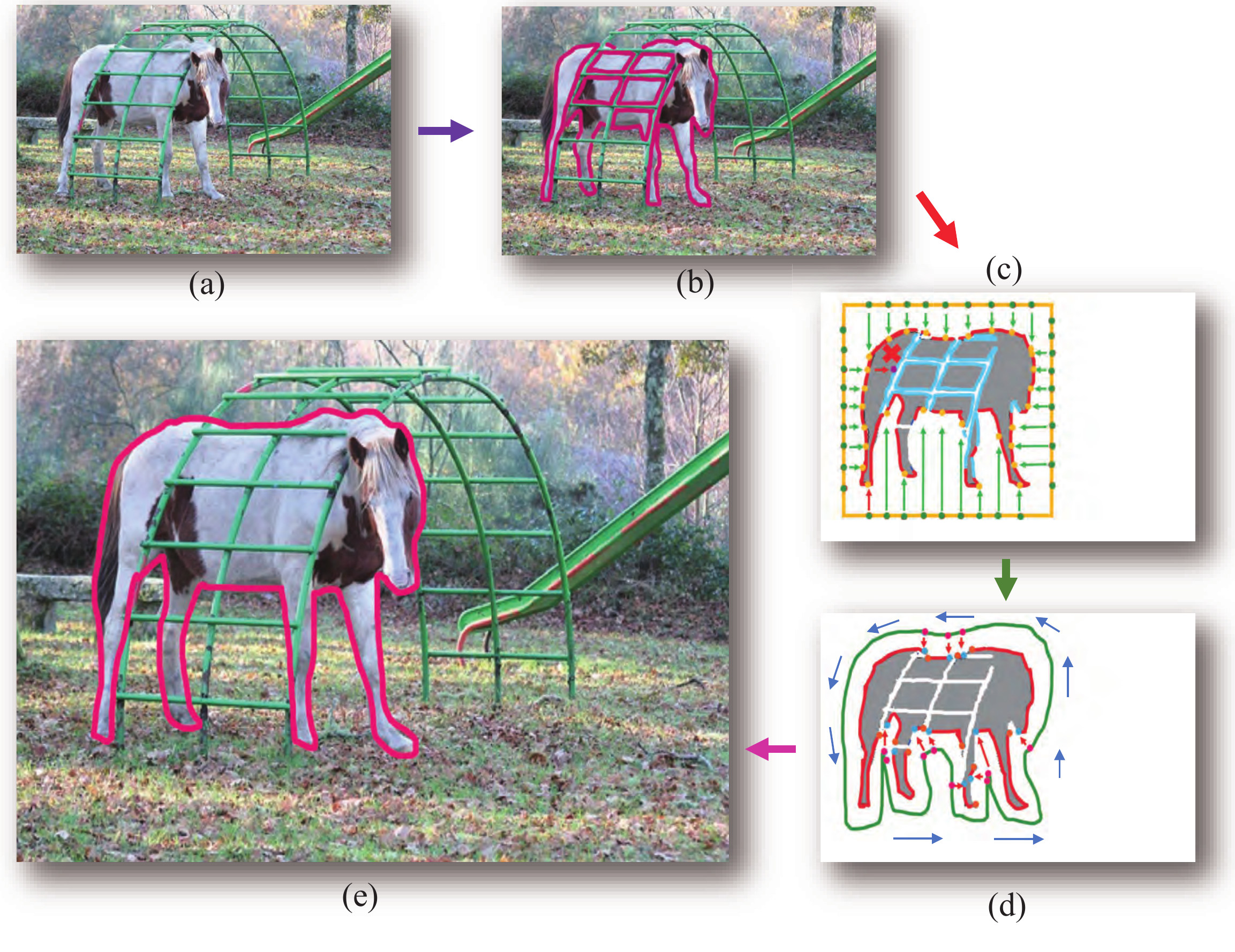}
\end{center}
   \caption{The process to complete the separated areas. An occluded object (a) has many separated areas (b), we split the contour points of each area into outline and inner points with the help of the bounding box (c), then we dilate the broken area into one, and reorder the outline points according to the dilated shape contour (d), finally, we complete the instance (e).}
\label{fig:dilate}
\end{figure}

In a few cases, an object is separated into disconnected regions, resulting in multiple inner centers. To deal with such situations, we dilate the broken areas to a single one and then find the contour of the dilated shape. Of course the contour is very rough, however, it can help to reorder the contour points of the outline points. The whole process is depicted in Fig. \ref{fig:dilate}. Thus inner center is computed from the completed contour. 

\paragraph{Dense Contour Sampling} We sample the contour points according to the angles at the interval of $\tau$ around inner-center point, thus a contour will result in $N=[2\pi/\tau]$ points. In practice, $\tau=\pi/180$ and thus $N=360$ points are sampled from an object contour. If the ray casting from the inner-center intersects more than once to the contour. We collect the point with the largest radius only. The function $\bm{f}(\theta)$ is denoted as radius at different angles $\theta$. To note, we are aware that the contour sampling in this way will not be perfect, however, after extensive experiments in Pascal VOC, and COCO, we find it suitable for natural objects (see Table \ref{exp:obj_sig}). A further discussion is in the next Sec. \ref{sec:dis}.

\subsubsection{Fitting the Signature to Coefficients}
\label{sec:fitting}
The IR makes shape representation into a vector. But, it is still too long for the network to train. Besides, the shape signature is very sensitive to the noise (see Fig. \ref{fig:sensitive_coco}). Thus, we take a further step to shorten shape vector and resist noise through Chebyshev polynomial fitting.

\paragraph{Chebyshev polynomials} The Chebyshev polynomial is defined in recurrence:
\begin{align}
   T_0(x)=1,\\T_1(x)=x,\\T_{n+1}(x)=2xT_n(x)-T_{n-1}(x),
\end{align}
which is also known as \textbf{The Chebyshev polynomials of the first kind}. It can effectively minimize the problem of \textit{Runge's phenomenon} and  provides a near-optimal approximation under the maximum norm\footnote{https://en.wikipedia.org/wiki/Chebyshev\_polynomials}.

Given the IR shape signature, the Chebyshev approximation is to find the coefficients in
$$\bm{f}(\theta)\sim \sum_{i=0}^\infty c_iT_i(\theta)$$
Truncating the function with $n$ terms, we have the approximation function $\tilde{f}(\theta)=\sum_{i=0}^n c_iT_i(\theta)$. $\bm{k}=(c_0,\ldots, c_n)$ are the shape signature vector to represent the object.

\subsubsection{Discussion}\label{sec:dis}
\begin{figure}[ht!]
\begin{center}
   \includegraphics[width=1\linewidth]{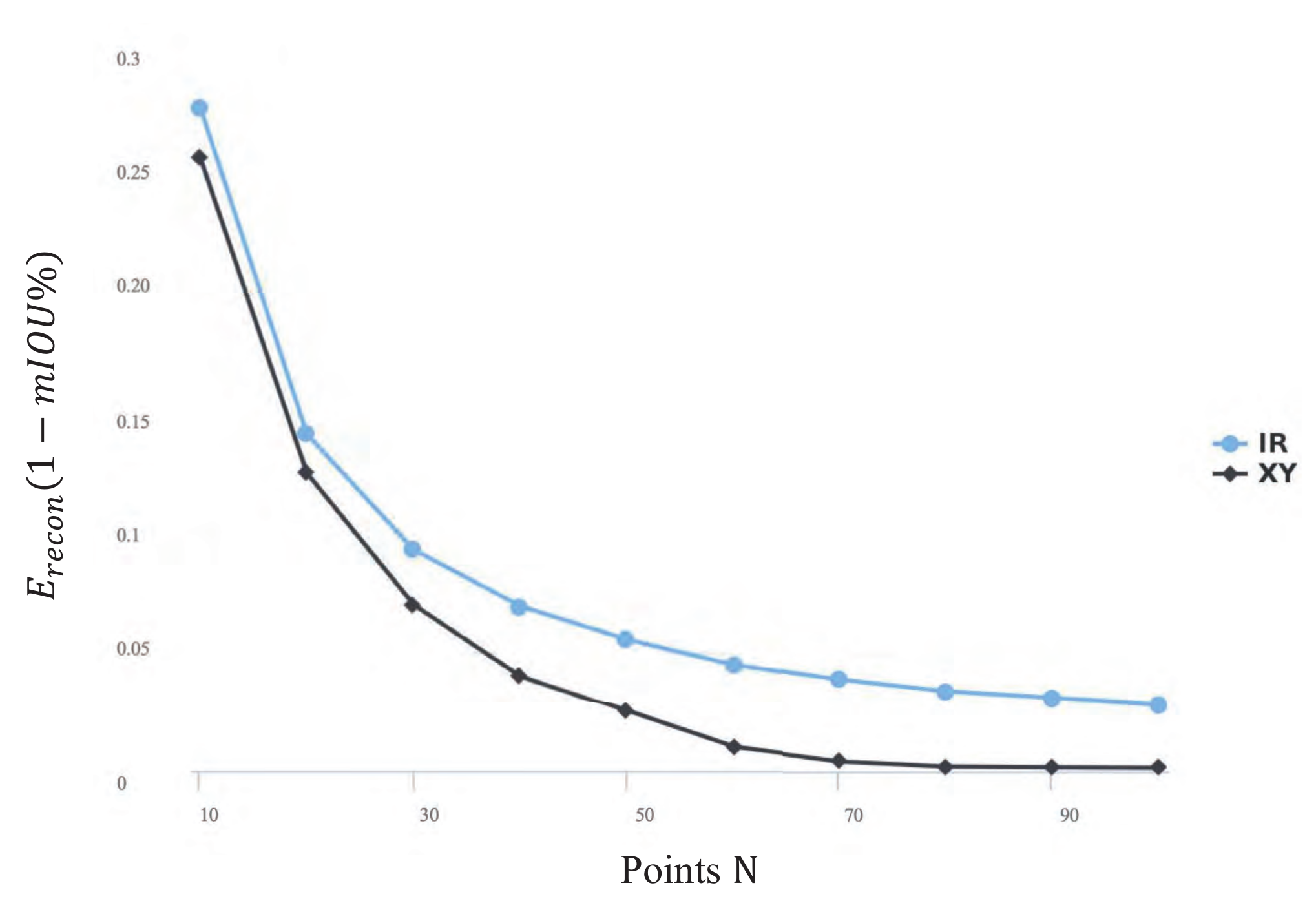}
\end{center}
   \caption{The reconstruction error $E_{recon}$ of IR and XY with different sampling number points.}
\label{fig:sig_recon_error}
\end{figure}

\paragraph{Comparison with Other Shape Signatures} The angle-based sampling for shape signature such as proposed IR is rarely adopted before, because it cannot perfectly fit shape segment. Actually, we compare and in-depth analyze other shape signatures and finally choose this solution. For example, a quite straight-forward design is to sample along the contour. The contour is represented by a set of contour polygon vertex coordinates. This method can nearly perfectly fit the object segment, especially non-convex shape. However, we find the performance of this design drops about $10$ mAP and more results are reported in Table \ref{exp:obj_sig}. The possible reason is that our angle-based sampling produces 1D sample sequence, yet, contour vertices sequence is a 2D sample sequence which is more sensitive to noise. We report the reconstruction error of these two shape signatures on Pascal VOC 2012 training in Fig. \ref{fig:sig_recon_error} (denoted as ``IR'' and ``XY'' respectively). Admittedly, the XY has less reconstruction error when sampling the same points on the contour, but when compared with the same dimension of the vector, IR is more accurate. For example, the dimension of the vector of IR at $N=20$ is the same as XY at $N=10$, the IR has a significantly less reconstruction error. Though when the $N$ gets larger, the difference gets smaller, a large $N$ will make training unstable as presented in Table \ref{exp:obj_sig}.

Other classic shape signatures such as centroid radius, cumulative angle cannot reconstruct the shape.

\paragraph{Comparison with Other Fitting Methods} Other commonly used function approximation methods, namely polynomial regression and Fourier series fitting are also considered.

For polynomial regression, the goal is to fit shape vector $\bm{k}=(v_0,\ldots,v_n)$, which is the coefficients of $n$ degree polynomials, $\tilde{\bm{f}}(\theta)=\sum_{i=0}^nv_ix^i$. For Fourier series fitting, the shape vector is $\bm{k}=(\omega, a_0, a_1,\ldots, a_n, b_1,\ldots,b_n)$, the truncated $n$ degree Fourier series is $\tilde{\bm{f}}(\theta)=a_0/2+\sum_{i=1}^n[a_i\cos(i\omega \theta)+b_i\sin(i\omega \theta)]$. As the dimension of $\bm{k}$ can be determined in advance, denoted as $l$. Thus we compare the methods from three aspects, \ie the reconstruction error $E_{recon}$, sensitivity to the noises, and the numeric distribution of the coefficients.

\begin{figure}[ht!]
\begin{center}
   \includegraphics[width=0.8\linewidth]{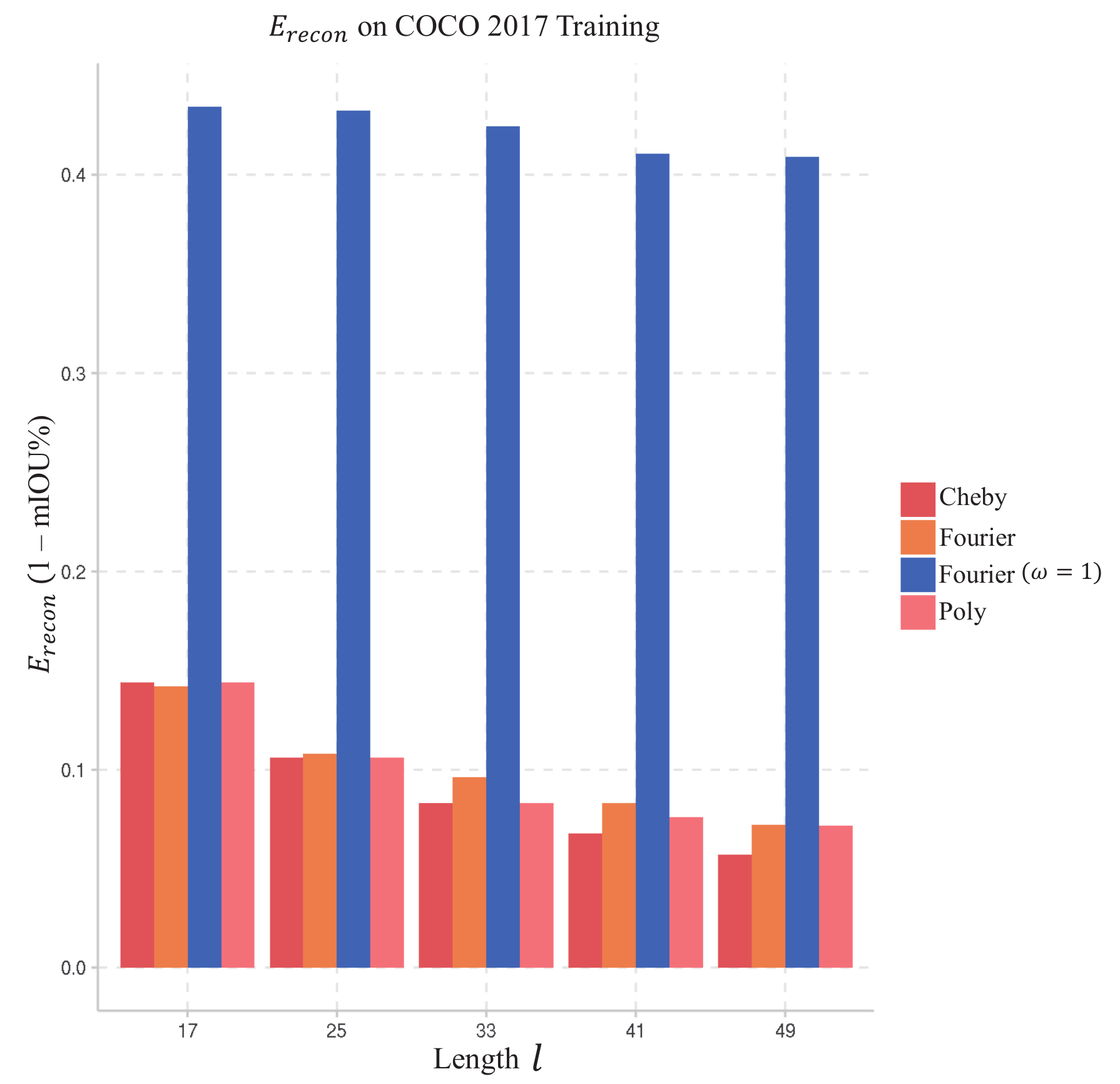}
\end{center}
   \caption{Comparison of $E_{recon}$ on COCO 2017 training.}\label{fig:recon_erro_coco}
\end{figure}

\begin{figure}[ht!]
\begin{center}
   \includegraphics[width=0.8\linewidth]{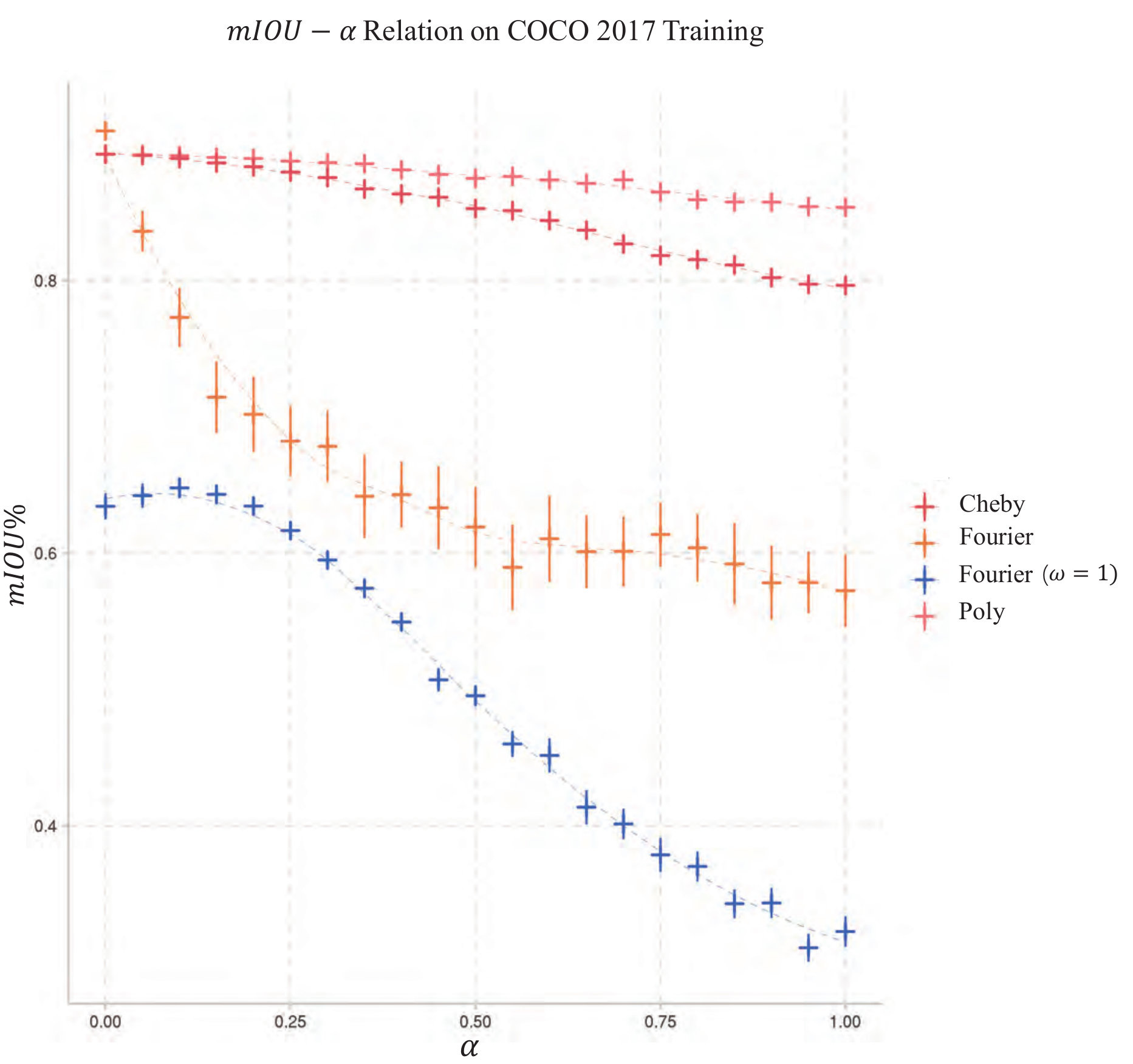}
\end{center}
   \caption{Comparison of the sensitivity on COCO 2017 training.}\label{fig:sensitive_coco}
\end{figure}

\begin{figure}[ht!]
   \begin{center}
      \includegraphics[width=0.8\linewidth]{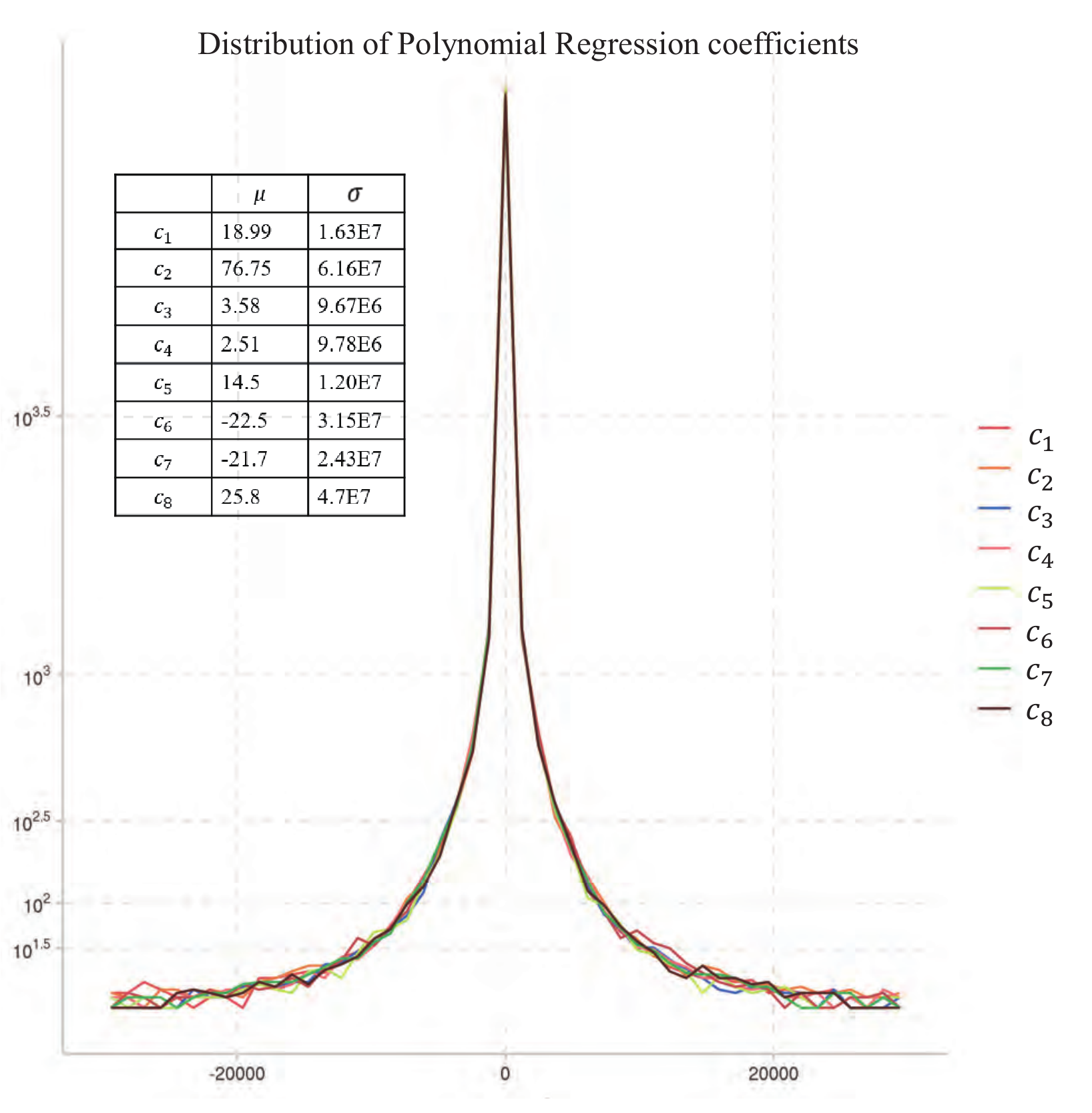}
   \end{center}
   \caption{The overall mean of the coefficients is , and the variance is for Polynomial regression.}\label{fig:coef_dist_poly}
\end{figure}

\begin{figure}[ht!]
   \begin{center}
      \includegraphics[width=0.8\linewidth]{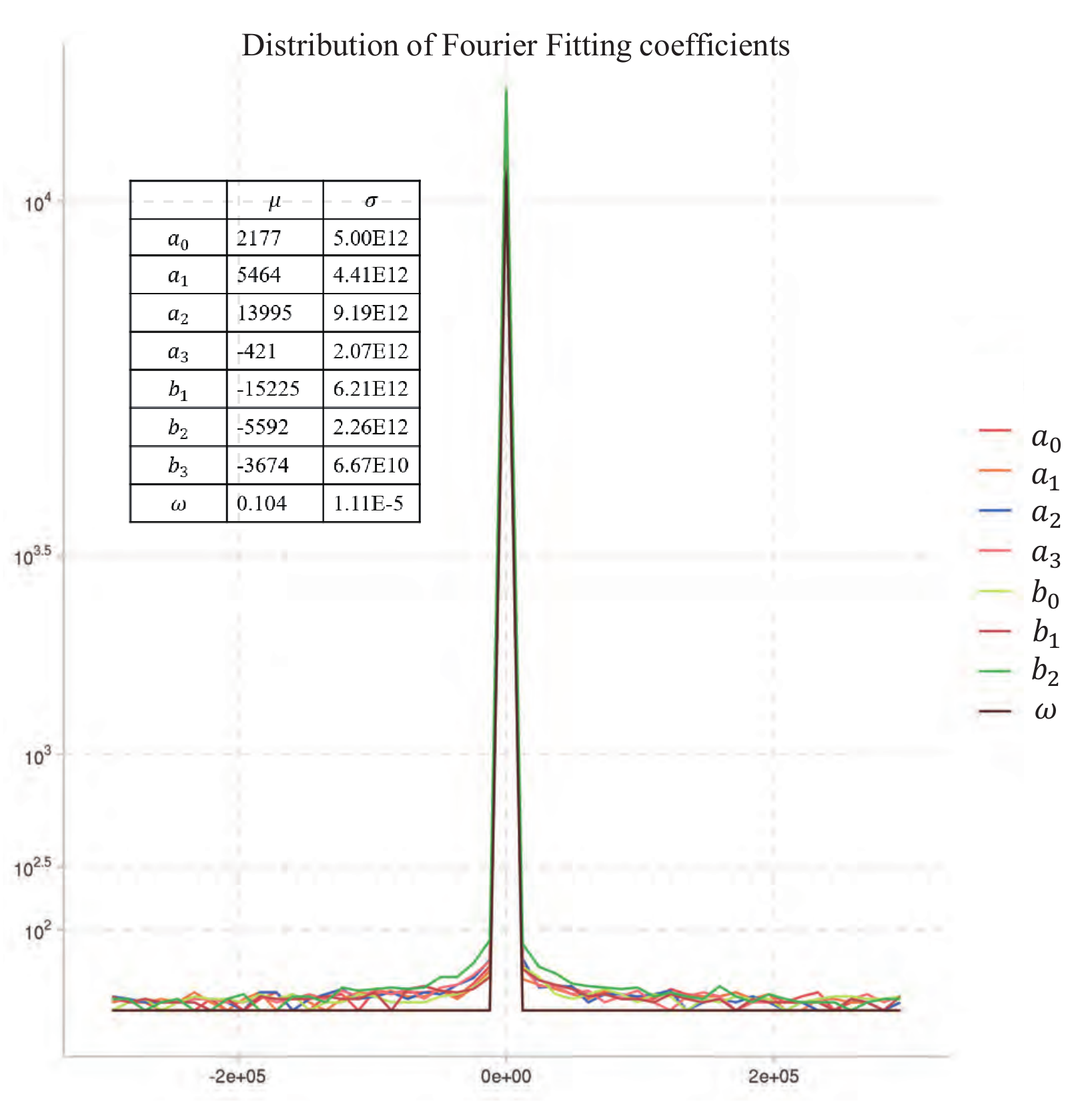}
   \end{center}
   \caption{Coefficients distribution of Fourier series fitting on COCO training 2017.}\label{fig:coef_dist_fourier}
\end{figure}

\begin{figure}[ht!]
   \begin{center}
      \includegraphics[width=0.8\linewidth]{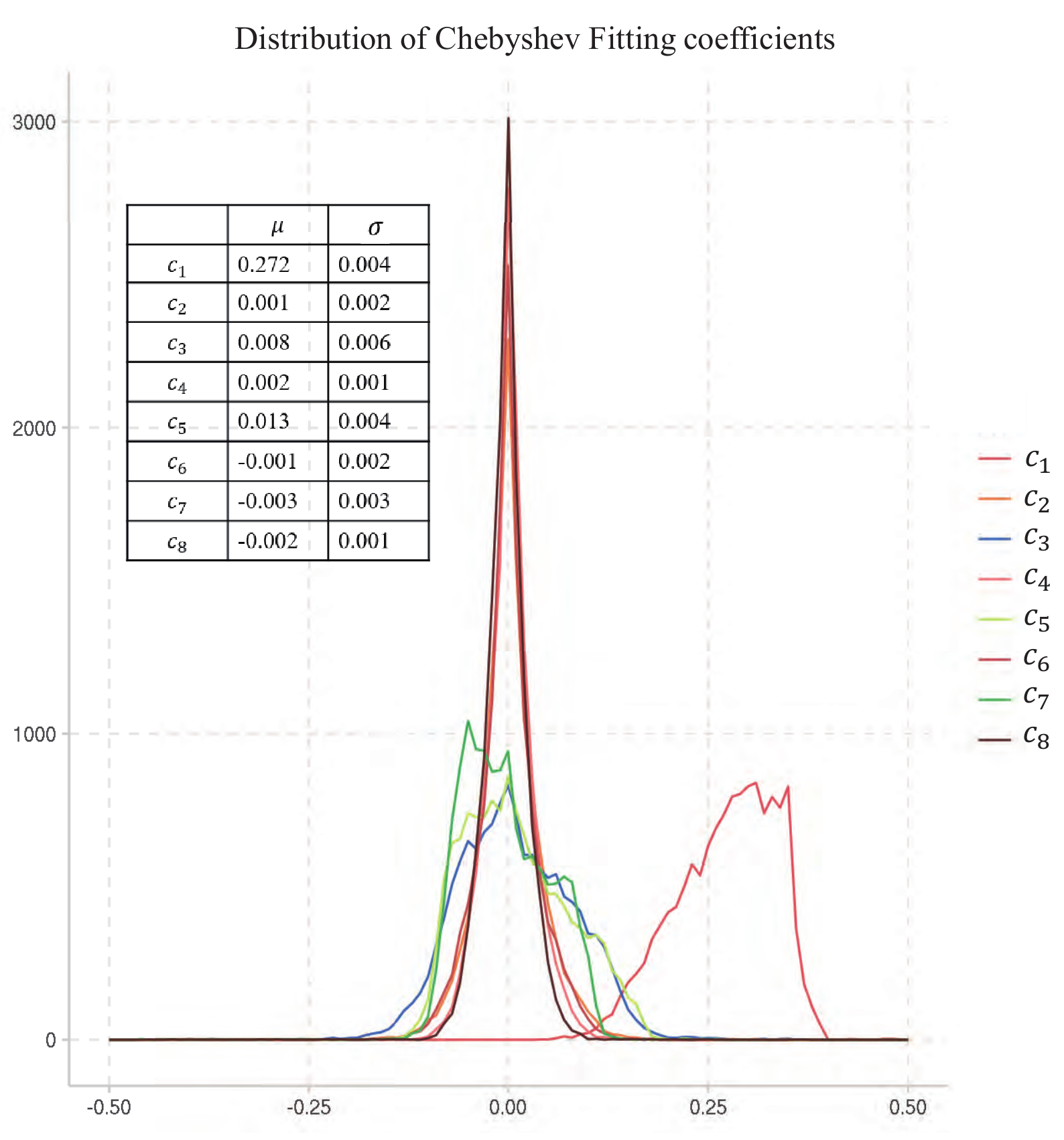}
   \end{center}
   \caption{Coefficients distribution of Chebyshev polynomial fitting on COCO training 2017.}\label{fig:coef_dist_cheby}
\end{figure}

The reconstruction errors $E_{recon}$ is calculated by $1-mIOU$ under the same dimension $l$ and point number $N$ in Fig. \ref{fig:recon_erro_coco}. Then we set $l = 8$ as an example to conduct the sensitivity analysis as shown in Fig. \ref{fig:sensitive_coco}. For each coefficient, it is interrupted by the noise $\varepsilon\sim N(0, \alpha\bar{k})$, $\bar{k}$ is the mean of the corresponding coefficient. As we can see, the $\omega$ of Fourier series is extremely sensitive, which may cause the Fourier fitting not suitable for the CNN training, as the CNN is known to regression with uncertainties. If we fix $\omega=1$, it becomes less sensitive, but has considerably larger reconstruction error. Besides, considering the difficulty for the network to learn, we also investigate the statistic on the distribution of the fitted coefficients. See Fig. \ref{fig:coef_dist_poly}, Fig. \ref{fig:coef_dist_fourier} and Fig. \ref{fig:coef_dist_cheby}. Chebyshev polynomials are better for shape signature fitting as it has less reconstruction error, less sensitivity to noise, better numeric distribution of coefficients.



\subsection{Regression Under Object Detection Framework}
\label{sec:learning}
Our network will learn to predict the inner center $\hat{\bm{p}}$, the shape vector $\hat{\bm{k}}$, along with the object bounding box. The loss function for bounding boxes regression, classification stays the same to the original object detection frameworks. For YOLOv3, the loss function for bounding box $\mathcal{L}_{bbox}$ and classification $\mathcal{L}_{cls}$ can be referred to \cite{yolov3}.
As for the loss function for the shape learning:
$$\mathcal{L}_{shape}=\mathbb{1}^{obj}||(\hat{\bm{p}}-\bm{p})+(\hat{\bm{k}}-\bm{k})||_2^2,$$
where $\mathbb{1}^{obj}$ indicates the grid cells with objects for the one-stage detectors, and the proposals for the two-stage detectors.
Thus the overall objective function is:
$$\mathcal{L} = \lambda_{cls}\mathcal{L}_{cls}+\lambda_{bbox}\mathcal{L}_{bbox} + \lambda_{shape}\mathcal{L}_{shape}$$

\subsection{Decoding Shape Vector to Shape}
\label{sec:decode}
Given the shape vector dimension $l$, the predicted shape vector $\hat{\bm{k}}=(\hat{k}_0,\ldots,\hat{k}_{l-1})^\top$, the fitted Chebyshev polynomial is $\hat{f}(\theta)=\sum_{i=0}^{l-1}\hat{k}_iT_i(\theta)$. And the polar coordinate transform factor $\bm{u}(\theta)=(\cos\theta, \sin\theta)$.Thus the shape can be recovered by traversing the $\theta \in [0, 2\pi)$
$$\hat{\bm{p}}_i=\hat{\bm{p}_c} + \hat{\bm{f}}(\theta)\odot\bm{u}(\theta).$$
$\odot$ is the Hadamard product. This calculation can be written in tensor operation form. Given the batch size $bs$, the corresponding tensor version are $\bm{\Theta}\in \mathbb{R}^{bs \times 1 \times N}$ for angles sampled, $\bm{\hat{C}} \in \mathbb{R}^{bs\times 1\times l}$ for the predicted shape vector, $\bm{\hat{P}_c} \in \mathbb{R}^{bs\times 2\times N}$ for the predicted inner centers and $\bm{\hat{P}}\in \mathbb{R}^{bs\times 2\times N}$ represents the decoded contour points. As expressed:  $$\bm{\hat{P}}=\bm{\hat{P}}_c + \bm{\hat{C}} T(\bm{\Theta})\odot\bm{u}^\top(\bm{\Theta}).$$
In the GPU setting, the computation cost of such tensor operation is very minor. Due to this extremely fast shape decoding, our instance segmentation can achieve the same speed with object detection. 

\section{Experiment}
We conduct extensive experiments to justify the descriptor choice and the efficacy of proposed methods. If not specified, the base detector is YOLOv3 implemented by GluonCV \cite{he2018bag_gluoncv}, the input image is $416\times 416$. $\lambda_{cls}=\lambda_{bbox}=\lambda_{shape}=1$. Other hyper-parameters stays the same as the YOLOv3 implementation. We trained 300 epochs and report the performance with the best evaluation results. For the model name with a bracket and a number in it, the number is the dimension of the shape vector.

\subsection{Explicit v.s. Implicit}
We first compare the explicit shape encoding with the implicit shape encoding. As the previous work \cite{wang_straight_to_shape_realtime_detect} provides a baseline for implicit shape representation with YOLO \cite{yolo} as the base detector, to be fairly compared, we also trained the ESE-Seg with YOLO base detector, the dimension of the shape vector is also the same. We denote the model as ``YOLO-Cheby (50)'' and ``YOLO-Cheby (20)''. The experiments are on Pascal SBD 2012 val \cite{BharathICCV2011}.

To note, the mainstream instance segmentation based on mask, namely SDS \cite{hariharan2014simultaneous}, MNC \cite{dai2016instance}, FCIS \cite{li2017fully}, Mask RCNN \cite{he2017maskrcnn}, can also be viewed as implicit shape encoding. We compare them with ``YOLOv3-Cheby (20)'' on Pascal VOC 2012 without SBD and COCO with their reported scores, which outperforms the Mask R-CNN (with ResNet50) at mAP$^r$@0.5 on Pascal VOC and close to it on COCO. To note, the input image size is 800 on the shorter side for Mask R-CNN with ResNet50-FPN, which is almost 4 times to our $416\times 416$. All results are reported in Table \ref{exp:comparison}.
\begin{table}[h]
\small
\centering
\begin{tabular}{c|c|c|c|c}
\toprule
\multicolumn{5}{c}{\textbf{SBD (5732 val images)}} \\
\hline
\hline
\diagbox{model}{mAP$^r$} &  0.5 & 0.7  & vol& Time (ms)\\
\midrule
BinaryMask\cite{wang_straight_to_shape_realtime_detect}& 32.3 & 12.0 & 28.6& 26.3  \\
Radial\cite{wang_straight_to_shape_realtime_detect}& 30.0 & 6.5  & 29.0 & 27.1 \\
Embedding (50) \cite{wang_straight_to_shape_realtime_detect} & 32.6 & 14.8  &28.9 & 30.5  \\
Embedding (20) \cite{wang_straight_to_shape_realtime_detect} & 34.6 & \textbf{15.0}  &31.5& 28.0 \\
\midrule
YOLO-Cheby (50) & 39.1 & 10.5 &32.6& 24.2 \\
YOLO-Cheby (20) & \textbf{40.7} & 12.1 &35.3& \textbf{24.0} \\
\hline
\multicolumn{5}{c}{\textbf{Pascal VOC 2012 val}}\\
\hline
\hline
\diagbox{model}{mAP$^r$} &  0.5 & 0.7 & vol& Time (ms)\\
\midrule
SDS  & 49.7 & 25.3 &41.4  & 48k \\
MNC  & 59.1 & 36.0 & - & 360 \\
FCIS  & 65.7 & \textbf{52.1} & - & 160\\
Mask R-CNN & 68.5 & 40.2 &-& 180\\
\midrule
YOLOv3-Cheby (20) & 62.6 & 32.4 &52.0& \textbf{26.0} \\
+ COCO pretrained & \textbf{69.3} & 36.7 &54.2& \textbf{26.0} \\
\hline
\multicolumn{5}{c}{\textbf{COCO 2017 val}}\\
\hline
\hline
\diagbox{model}{mAP} &  0.5 & 0.75 &all& Time (ms)\\
\midrule
FCIS & 49.5 & -  &29.2& 160\\
Mask R-CNN & 51.2 & 31.5  &30.3& 180\\
\midrule
YOLOv3-Cheby (20) & 48.7 & 22.4 &21.6& \textbf{26.0}\\
\bottomrule
\end{tabular}
\caption{Comparison of ESE-Seg to the previous methods on Pascal SBD 2012 val, Pascal VOC 2012 without SBD val, and COCO 2017 val.}\label{exp:comparison}
\end{table}
\begin{figure*}[ht!]
\begin{center}
   \includegraphics[width=0.9\linewidth]{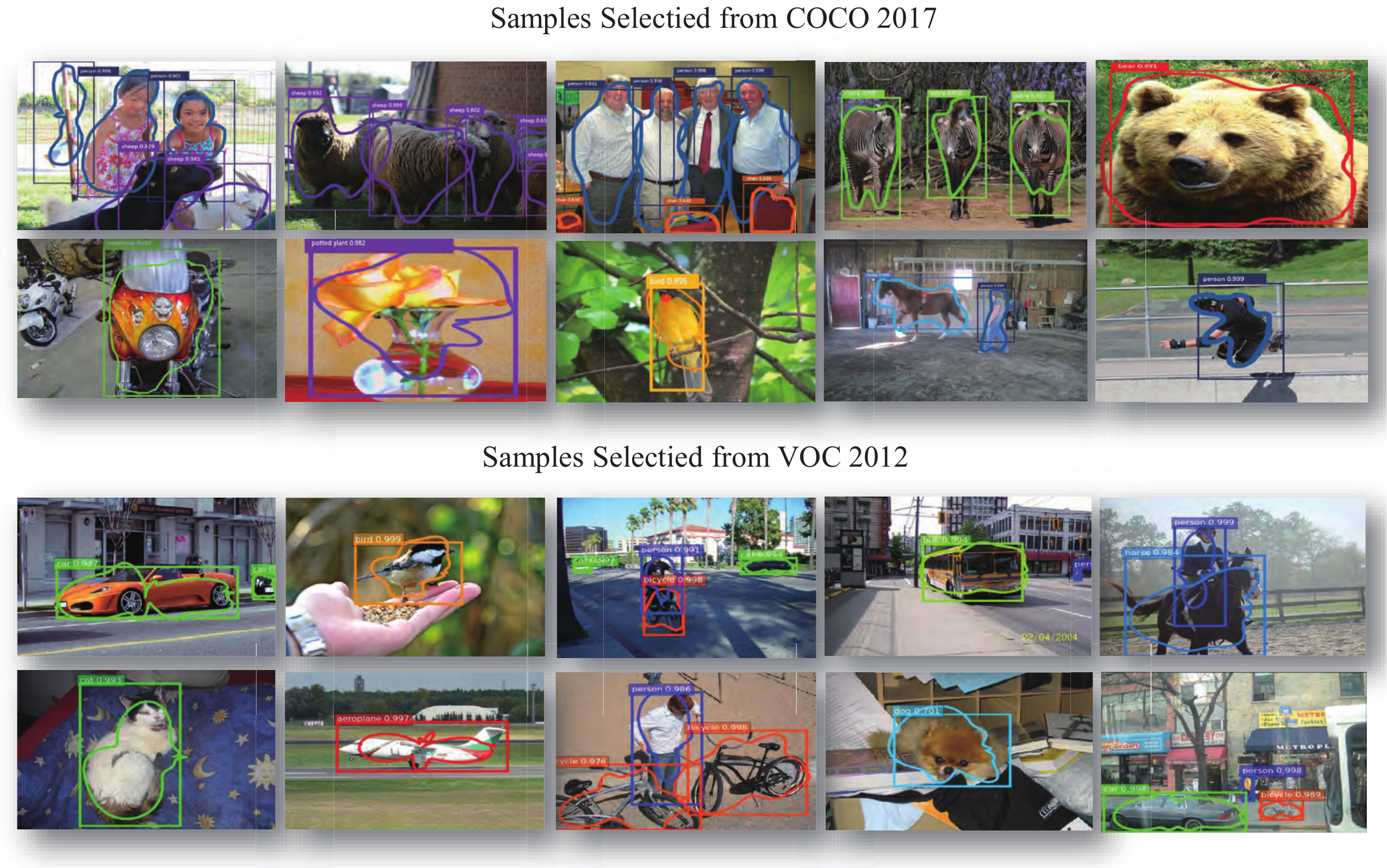}
\end{center}
   \caption{Qualitative results generated by our methods.}
\label{fig:quality}
\end{figure*}

\subsection{On explicit descriptors}
In this section, we will compare the object shape signatures and the function approximation methods quantitatively.
\paragraph{On Different Shape Signatures}
For object shape signatures, we compare our proposed IR with a straightforward 2D vertices representation on Pascal VOC 2012. (See Table \ref{exp:obj_sig}) We adopt the squared boxes, \ie the bounding box, as the baseline. To note, the squared boxes baseline is not the object detection scores, as the baseline computes the IoU between the bounding box and the instance mask.

For each shape signature, we compare regressing directly and regressing after Chebyshev polynomial fitting. For direct regression, we control the length of the shape signature by adjusting the $\tau$ for each shape. We select $20$ and $40$ points to regress. We denote model trained on 2D vertices ``XY'', the shape vector has a dimension of $40$ and $80$ respectively. As for the Chebyshev fitting on these signatures, we fit the $x$ coordinates and $y$ coordinates respectively. Denoted as ``XY-Cheby (10+10)'' means each fitted function has 10 coefficients.

\begin{table}[h]
\small
\centering
\begin{tabular}{c|c|c}
\toprule
\diagbox{model}{mAP$^r$} &  0.5 & 0.7 \\
\midrule
Squared Boxes & 42.3 & 8.6  \\
\midrule
XY (20) & 46.1 & 10.7 \\
XY (40) & 43.5& 11.2\\
XY-Cheby (10+10) & 48.3 & 16.4 \\
XY-Cheby (20+20) & 53.1 & 20.9 \\
\midrule
IR (20) & 48.8& 13.5\\
IR (40) & 52.6& 19.3 \\
IR (60) & 51.7& 16.4 \\
IR-Cheby (20) & 62.6 & 32.4 \\
\bottomrule
\end{tabular}
\caption{We compare different choice of the shape signatures on Pascal VOC 2012.}\label{exp:obj_sig}
\end{table}

\paragraph{On Different Function Approximation Techniques}
We have already compared the function approximation techniques through off-line analysis. However, it is still interesting to know performance of the neural network on the coefficients obtained by these methods.

All the function approximations are carried out on IR $\bm{f}(\theta)$. The polynomial regression is denoted as ``Poly'', while ``Fourier'' for Fourier series fitting and ``Cheby'' for Chebyshev polynomial fitting. All models have tested on Pascal VOC 2012 val. See Table \ref{exp:obj_fitting}.
\begin{table}[h]
\small
\centering
\begin{tabular}{c|c|c}
\toprule
\diagbox{model}{mAP$^r$} &  0.5 & 0.7  \\
\midrule
Poly (20) & 26.3 & 5.4\\
Fourier (20) & 37.5 & 9.1 \\
Fourier (40) & 36.1 & 8.5\\
\midrule
Cheby (20) & 62.6 & 32.4 \\
Cheby (40) & 60.7 & 31.5 \\
\bottomrule
\end{tabular}
\caption{Comparison of the performance of different shape signatures on Pascal VOC 2012 val.}\label{exp:obj_fitting}
\end{table}

\subsection{On base object detector}
To show the generality of the object shape detection, we also conduct the shape learning on Faster R-CNN (``Faster-Cheby (20)''), RetinaNet (``Retina-Cheby (20)'') and YOLOv3-tiny (``YOLOv3-tiny-Cheby (20)''). Not only the performance is stable for all these bounding box-based detectors, but the speed boost due to the detector can be enjoyed. As shown in Table \ref{exp:base}.
\begin{table}[h]
\small
\centering
\begin{tabular}{c|c|c|c|c}
\toprule
\diagbox{model}{mAP$^r$} & 0.5 & 0.7 &vol&  Tims (ms) \\
\midrule
 YOLOv3-Cheby (20) & 62.6 & 32.4 &52.0& 26  \\
 Faster-Cheby (20) & 63.4  & 32.8 &54.2& 180  \\
 Retina-Cheby (20)  & 65.9 & 36.5&56.7& 73 \\
 YOLOv3-tiny-Cheby (20) & 53.2 & 15.8 &42.5& \textbf{8}\\
\bottomrule
\end{tabular}
\caption{Comparison of different base object detectors with IR shape signature and Chebyshev fitting on Pascal VOC 2012 val.}\label{exp:base}
\end{table}
\subsection{Qualitative Results}
Qualitative results are shown in Fig. \ref{fig:quality}. Obviously, the predicted shape vectors indeed capture the characteristics of the contours, not produce the random noise.


\section{Limitations and Future Works}
Our proposed ESE-Seg can achieve the instance segmentation with minor time-consumption, with a decent performance at IoU threshold 0.5. However, due to the inaccuracy $E_{recon}$ of the shape vector, and the noise comes with the CNN regression, performance at larger IoU threshold like 0.7 drop a large margin. In the future, better ways to explicitly represent the shape, and better ways to train the CNN regression which will contribute to higher performance at high IOU threshold are of high interest.

\paragraph{Acknowledgement} This work is supported in part by the National Key R\&D Program of China, No. 2017YFA0700800, National Natural Science Foundation of China under Grants 61772332.

{\small
\bibliographystyle{ieee_fullname}
\bibliography{shape}
}

\end{document}